\documentclass{gretsi}
\usepackage[utf8]{inputenc}
\usepackage[T1]{fontenc}
 \usepackage[english,french]{babel}   
\usepackage{graphicx}
\usepackage{float}
\usepackage{amsmath}
\usepackage{amsthm}
\usepackage{amssymb}
\usepackage[font=small]{caption}
\usepackage{subcaption}
\usepackage{xcolor}
\titre{ScoreCAM GNN : une explication optimale des réseaux profonds sur graphes}
\auteur{\coord{Adrien}{Raison}{1,2}, \coord{Pascal}{Bourdon}{1,2}, \coord{David}{Helbert}{1,2}}
\adresse{\affil{1}{XLIM-Asali, UMR CNRS 7252, Université de Poitiers}\affil{2}{I3M Laboratoire Commun CNRS-Siemens, Université de Poitiers}}
\email{prénom.nom@univ-poitiers.fr}
\resumefrancais{L'explicabilité des réseaux profonds devient un problème central dans la communauté de l'apprentissage profond. Il en est de même pour l'apprentissage sur graphes, structure de données présente dans bon nombre de problématiques du monde réel. Nous proposons dans cette contribution, une méthode plus optimale, plus légère, consistante et exploitant mieux la topologie des graphes évalués que les méthodes de l'état de l'art.}
\resumeanglais{The explainability of deep networks is becoming a central issue in the deep learning community. It is the same for learning on graphs, data structure present in many problems of the real world. In this paper, we propose a method that is more optimal, lighter, consitent and exploits the of the evaluated graph that the methods of the state of the art.}
\newtheorem{defn}{Définition}[section]

\newcommand\widthpic{0.16}
\newcommand\hsp{0.05cm}
\begin{document}
\maketitle
\section{Introduction}
Les réseaux profonds sont largement utilisés aujourd'hui pour résoudre des problèmes complexes de nature académiques, industriels ou encore sociaux, dus à leur performance. À cause de la dimension de l'espace des paramètres ou encore de l'architecture des réseaux, les canaux d'informations constitués au cours de l'apprentissage sont compliqués à appréhender humainement. Il y a une difficulté à rendre intelligible le lien entre données d'entrée et de sortie, qui eux, sont parfaitement intelligibles, interprétables et compréhensibles. Ces architectures sont souvent qualifiées de ``boites noires'' dans la littérature. Le domaine de l'apprentissage profond sur graphe subit aussi ces inconvénients. Des méthodes s'intéressant à résoudre ces problèmes d'explicabilité des réseaux profonds sur graphes ont été proposées dernièrement. Cependant, l'interprétation des cartes de saillance est délicate car dépendante du contexte. Ces cartes doivent donc être soumises à expertise. Pour s'affranchir de cela et aussi du caractère subjectif des expertises, des métriques objectives ont été introduites \cite{yeh_delity_nodate}\cite{adebayo_sanity_nodate}. La construction des explications d'une instance revient souvent à filtrer une partie de celle-ci pour extraire que l'information pertinente par rapport à la tâche d'apprentissage. Dans le cas d'une structure de donnée comme les graphes, il s'agit d’échantillonner intelligemment des sous-graphes, un problème d'optimisation souvent de type combinatoire. Dans cette étude, nous proposons une méthode plus légère, plus optimale, et plus consistante que celles proposées par l'état de l'art. Nous caractériserons d'une manière objective et subjective la pertinence de notre méthode par rapport à l'état de l'art.
Dans le contexte d'apprentissage profond sur graphes, il est commun de considérer un graphe $G$ à $N$ noeuds comme étant un couple ($\mathbf{X},\mathbf{A}$) où $\mathbf{X}\in\mathbb{R}^{N\times F}$ est la matrice des descripteurs (de dimension $F$) des noeuds et où $\mathbf{A}\in\{0,1\}^{N\times N}$ est la matrice d'adjacence de $G$. Les couches d'apprentissage profond sur graphes sont structurées de manière récursive. Pour un module de profondeur $L$, on a pour tout $l\in\{1,\dots,L\}$,$\mathbf{H^{(l)}}=\sigma(\mathbf{A}\mathbf{H^{(l-1)}}\mathbf{W^{(l-1)}})$ avec $\mathbf{H}^{(0)}=\mathbf{X}$, $\mathbf{W^{(l-1)}}$ est la matrice des paramètres de la $(l-1)$-ème couche et où $\sigma$ est une fonction d'activation. Pour tenter de répondre aux problématiques d’explicabilité, plusieurs méthodes ont été proposées. XGNN \cite{yuan_xgnn_2020} construit ses cartes de saillance par un processus décisionnel de Markov apprenant une politique optimale. Cette méthode, sensible aux conditions d'initiales,  nécessite l'entrainement d'un modèle supplémentaire (i.e en plus du classifieur) assurant une solution optimale asymptotiquement, donc pas atteignable en pratique. SubgraphX \cite{yuan_explainability_nodate} sous échantillonne le graphe et optimise son procédé de construction des cartes de saillance au travers un arbre de recherche Monte-Carlo. L'importance du sous-graphe est donnée par la valeur \textit{Shapely}. Cette méthode est elle aussi par essence sensible aux conditions initiales et ne propose qu'une solution optimale asymptotiquement. Des méthodes comme GNNExplainer \cite{ying_gnnexplainer_nodate} ou encore PGExplainer \cite{luo_parameterized_nodate} s'intéressent à optimiser l'information mutuelle comprise entre des sous-graphes échantillonnés et la sortie du classifieur. GraphSVX \cite{duval2021graphsvx} est une méthode basé sur l'échantillonange de sous-graphes. Elle optimise un modèle linéaire dont les paramètres convergent vers les valeurs \textit{Shapely} de chaque sous-graphe.

L'inconvénient de recourir à l'échantillonnage de sous graphes est de construire des cartes de saillances basées sur un socle variable. Cela produit donc des explications différentes pour un même modèle et une même instance. Or les fondements d'une explication faite à propos d'un unique phénomène ne doivent pas varier. De plus, la plupart des méthodes propose des approximations de cartes de saillance théoriquement atteignables et il a un réel intérêt à avoir accès aux cartes optimales lorsque l'on est sur des problématiques sensibles, comme celles du domaine médical ou judiciaire. Dans cette étude, nous proposons une méthode ne nécessitant pas d'échantillonnage ni de post-entrainement et fournissant des explications de meilleure qualité, concises, non redondantes et concentrée sur la topologie des graphes.
\section{Score CAM GNN}
La méthode ScoreCAM \cite{wang_score-cam_2020} initialement développée pour les CNN, est dérivée de la méthode \textit{Class Activation Mapping} (CAM) \cite{zhou_learning_2016}. La méthode construit ses cartes de saillance en combinant linéairement les poids de la couche de moyennage d'agrégation globale aux couches de descripteurs de plus hauts niveaux. Un inconvénient majeur de cette méthode est d'imposer une architecture particulière au classifieur. La méthode GradCAM \cite{yuan_explainability_nodate} est une approche gradient basée sur le même principe qui s'affranchit de la nécessité d'avoir une architecture particulière, et donc qui est adaptable n'importe laquelle d'entre-elles. Elle souffre cependant du phénomène d'évaporation du gradient ou du principe de fausse attribution lui empêchant une totale expressivité. La méthode ScoreCAM est, elle, adaptable à n'importe quelle architecture et évite les problématiques liées au gradient. Par un procédé d'évaluation de carte de descripteurs masquée par l'entrée initiale, la méthode combine linéairement les scores de ces évaluations par rapport à la tâche de classification avec les couches de descripteurs de plus haut niveau pour construire ses cartes de saillance. Cette étude est porter sur l'extension de la méthode ScoreCAM à des données type graphe.
%Contrairement aux méthodes de l'état de l'art, cette extension permet d'obtenir de manière plus \textbf{optimale}, plus \textbf{lègere} à obtenir d'un point de vue computationnel, \textbf{consistente} les cartes de saillances sur graphes.
\begin{defn}
Pour un graphe $G=(\mathbf{X},\mathbf{A})$ et une paire de masques $(\mathbf{M_X},\mathbf{M_A})$, on définit :
  $$G\star(\mathbf{M_F},\mathbf{M_A}) = (\mathbf{X}\circ\mathbf{M_X},\mathbf{A}\circ\mathbf{M_A})$$
  où $\circ$ est le produit d'Hadamard.
\end{defn}
\begin{defn}
Pour $\textbf{x}\in\mathbb{R}^n$ tel que pour tout $\lambda\in\mathbb{R}$, $\textbf{x}\neq\lambda\mathbf{1}$, où $\mathbf{1}=(1,\dots,1)\in\mathbb{R}^n$. On définit :
  \begin{table}[h]
    \centering
  \begin{tabular}{llcl}
N: & $\mathbb{R}^n$ & $\rightarrow$ & $\mathbb{R}^n$\\
    & \textbf{x} & $\mapsto$ & $\frac{\textbf{x}-\underset{i}{\min \textbf{x}_i}}{\underset{i}{\max \textbf{x}_i}-\underset{i}{\min \textbf{x}_i}}$
  \end{tabular}
\end{table}
\end{defn}
Le graphe expliqué par la méthode ScoreCAM GNN est défini par :
\begin{defn}
  Pour un graphe $G=(\mathbf{X},\mathbf{A})$ et un classifieur $\phi_{\theta}$ de profondeur $L$ avec $\mathbf{H}^L=(\mathbf{h}_k^L)_k$ la dernière carte de descripteurs de $\phi_{\theta}$ on a :
  $$\psi_{SCGNN}(G,\phi_{\mathbf{\theta}})=(ReLU(\sum_k u_k \textbf{h}_k^L),\mathbf{A})$$
  où pour tout $k$, $$u_k=\frac{e^{\phi_{\mathbf{\theta}}(G\star(\textbf{N}(\textbf{h}_k^L),\mathbf{1}))_c}}{\sum_i e^{\phi_{\theta}(G\star(\textbf{N}(\textbf{h}_i^L),\mathbf{1}))_c}}$$
  et où $c=\arg\max \mathbf{\phi_{\mathbf{\theta}}}(G)$
\end{defn}
\begin{figure}[H]
  \centering
  \includegraphics[width=0.4\textwidth]{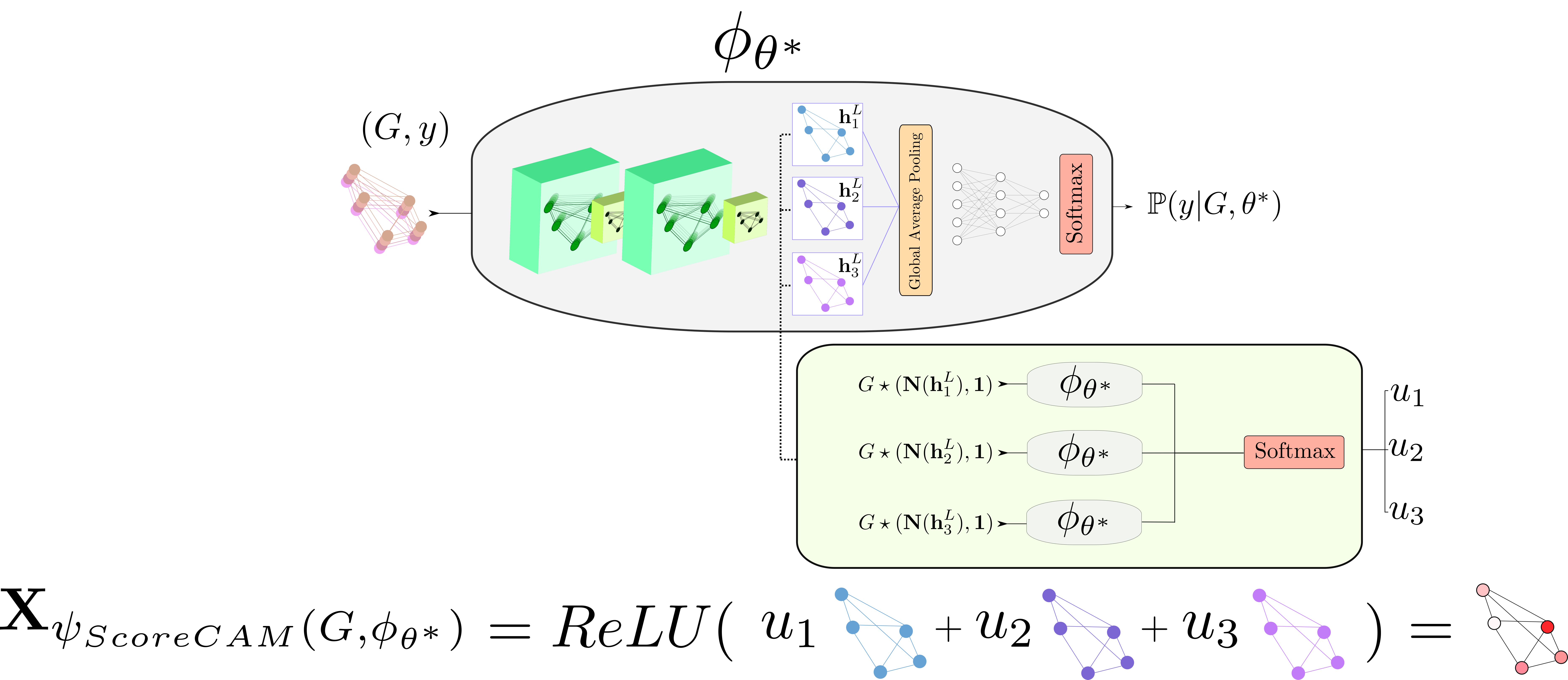}
  \caption{Configuration de la méthode ScoreCAM GNN}
  \label{fig:struc}
\end{figure}
\section{Résultats}
Nous avons évalué ScoreCAM GNN sur l'ensemble de données MNISTSuperpixels. Cet ensemble introduit par \cite{monti_geometric_2017} est basé sur l'ensemble de données MNIST \cite{lecun_gradient-based_1998}. Il contient 70 000 exemples répartis sur 10 classes. Pour chaque exemple est fourni un graphe où chaque nœud est un superpixel de l'instance image associée dans MNIST; il possède aussi la même étiquette. Les nœuds sont reliés en fonction de leurs relations spatiales. Pour classifier les instances de MNISTSuperpixels, nous avons utilisé dans deux réseaux différents, les deux modules d'apprentissage profonds les plus utilisés par l'état de l'art pour traiter des données type graphes : \textit{Graph Convolutional Network} (GCN) \cite{kipf_semi-supervised_2017} et \textit{Graph Attention Network} (GAT) \cite{velickovic_graph_2018}. Ils seront indiqués par ``Module Graphe'' ou (MG) dans la description suivante :
\begin{center}
  \small
  $(\mathbf{X},\mathbf{A})\rightarrow$ MG(1,64) $\rightarrow$ TanH $\rightarrow$ (MG(64,64) $\rightarrow$ TanH) $\times$ 3 $\rightarrow$ [\textit{Global Average Pooling},\textit{Global Max Pooling}] $\rightarrow$ Linéaire(128,10)  
\end{center}
Nous avons utilisé l'optimiseur Adam avec un taux d'apprentissage de $7\times 10^{-4}$. Nous avons obtenu 65\% de précisions au bout de 500 épochs.
%\pan{Ealuation axiomatique}
\paragraph{Évaluation objective}
Expliquer les prises de décisions effectuées par un classifieur profond est souvent lié à la nécessité d'évaluer la pertinence des cartes de saillance obtenue relativement au contexte. Cela requiert donc l'évaluation des ces dernières par un expert, elles sont par essence subjectives et peuvent être biaisées. Dans \cite{yeh_delity_nodate}, une métrique d'évaluation de la pertinence des cartes de saillances de manière objective est proposée. Elle quantifie la fidelité d'une méthode explicative par rapport a une instance et son classifieur.
Une métrique complémentaire utile pour évaluer la qualité de cartes de saillance est d'évaluer sa parcimonie. En effet, une bonne carte de saillance doit se concentrer avec minutie uniquement sur les points d'intérêt et ne pas attribuer d'importance aux autres.
Ces deux métriques sont de bons moyens pour évaluer la justesse globale d'une méthode d'explicabilité.
Pour le modèle basé sur GCN, ScoreCAM GNN est 89\% moins infidèle que le taux d'infidélité moyen observé par rapport aux autres méthodes. Elle atteint un taux de parcimonie 3.3 fois plus important que le taux moyen observé sur les autres méthodes.
Concernant le modèle GAT, ScoreCAM GNN est respectivement 93\% moins infidèle que le taux d'infidélité moyen observé par rapport aux autres méthodes. Sa parcimonie est 2.2 fois plus importante que la parcimonie moyenne observée sur les autres méthodes de l'état de l'art.
\begin{table}[H]
\centering  
\tiny
\begin{tabular}{|c|c|c|c|}
\hline
\textbf{XAI Method} & \textbf{Model type} & \textbf{Mean Infidelity} ($\downarrow$) & \textbf{Mean Sparsity} ($\uparrow$) \\ \hline
DL & GAT & 307.08 ($\pm$ 469.43) & 0.10 ($\pm$ 0.05) \\ \hline
GCAM & GAT & 85.84 ($\pm$ 62.94) & 0.48 ($\pm$ 0.15) \\ \hline
GNNEX & GAT & 201.67 ($\pm$ 106.18) & 0.51 ($\pm$ 0.38) \\ \hline
GSVX & GAT & 851.77 ($\pm$ 132.90) & 0.01 ($\pm$ 0.34) \\ \hline
\textbf{SCGNN} & \textbf{GAT} & \textbf{23.89($\pm$ 0.37)} & \textbf{0.62 ($\pm$ 0.08)} \\ \hline
DL & GCN & 862.51 ($\pm$ 26.31) & 0.14 ($\pm$ 0.06) \\ \hline
GCAM & GCN & 340.88 ($\pm$ 92.17) & 0.46 ($\pm$ 0.15) \\ \hline
GNNEX & GCN & 592.14 ($\pm$ 31.33) & 0.27 ($\pm$ 0.19) \\ \hline
GSVX & GCN & 583.58 ($\pm$ 23.15) & 0.01 ($\pm$ 0.03) \\ \hline
\textbf{SCGNN} & \textbf{GCN} & \textbf{68.67 ($\pm$ 7.01)} & \textbf{0.74($\pm$ 0.04)} \\ \hline
\end{tabular}
\caption{Métriques objectives sur MNISTSuperpixels}
  \label{perf_MS}
\end{table}
\paragraph{Évaluation subjective}
Pour établir visuellement la pertinence de la méthode ScoreCAM GNN, nous avons superposé, pour une instance donnée, son image et son graphe de superpixels associé. Classifier des chiffres non égaux entre eux demande de se concentrer sur leur topologie de manière à pouvoir établir des critères de différenciation pertinents. Expliquer comment classifier tel ou tel chiffre doit donc être basé sur des arguments structurels. De par sa définition, la méthode ScoreCAM GNN est proche de l'architecture du classifieur qu'elle explique. Son avantage par rapport à l'état de l'art est de fournir des explications non redondantes et prendre en compte pleinement la topologie du graphe.
\begin{figure*}
\centering
  \begin{subfigure}{\widthpic\textwidth}
    \includegraphics[width=\textwidth]{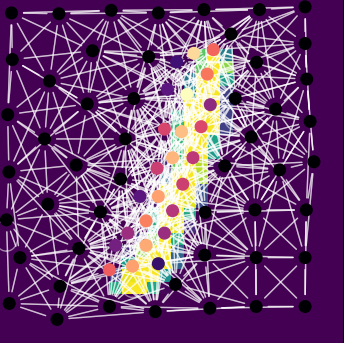}
    \caption{DeepLIFT-$n^{\circ}$346}
    \label{figvisu:DL346}
  \end{subfigure}\hspace{\hsp}%
\begin{subfigure}{\widthpic\textwidth}
    \includegraphics[width=\textwidth]{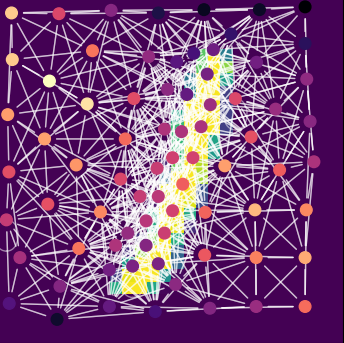}
    \caption{GradCAM-$n^{\circ}$346}
    \label{figvisu:GCAM346}
  \end{subfigure}\hspace{\hsp}%
%\begin{subfigure}{\widthpic\textwidth}
    %\includegraphics[width=\textwidth]{plt_MNIST_346_GNNEX.png}
    %\caption{\ GNNExplainer - $n^{\circ}$ 346}
    %\label{figvisu:GNNEX346}
  %\end{subfigure}\hspace{\hsp}%
 \begin{subfigure}{\widthpic\textwidth}
  \centering
    \includegraphics[width=\textwidth]{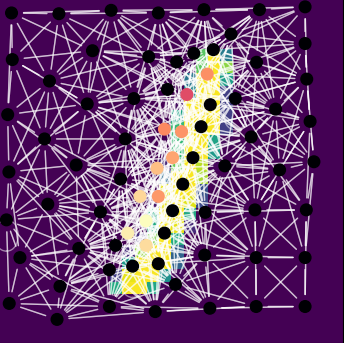}
    \caption{ScoreCAM-$n^{\circ}$346}
    \label{figvisu:SCGNN346}
  \end{subfigure}\hspace{\hsp}
   \begin{subfigure}{\widthpic\textwidth}
    \includegraphics[width=\textwidth]{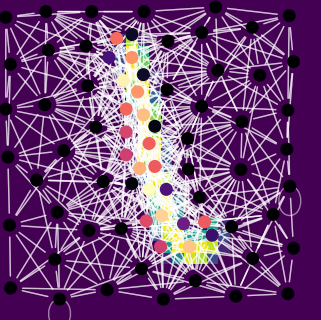}
    \caption{DeepLIFT-$n^{\circ}$478}
    \label{figvisu:DL478}
  \end{subfigure}\hspace{\hsp}%
\begin{subfigure}{\widthpic\textwidth}
    \includegraphics[width=\textwidth]{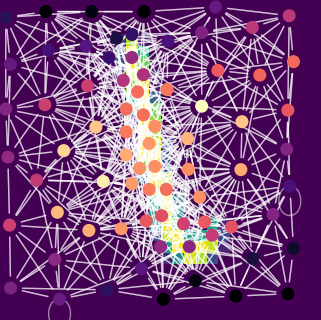}
    \caption{GradCAM-$n^{\circ}$478}
    \label{figvisu:GCAM478}
  \end{subfigure}\hspace{\hsp}%
%\begin{subfigure}{\widthpic\textwidth}
    %\includegraphics[width=\textwidth]{plt_MNIST_478_GNNEX.png}
    %\caption{\ GNNExplainer - $n^{\circ}$ 478}
    %\label{figvisu:GNNEX478}
  %\end{subfigure}\hspace{\hsp}%
\begin{subfigure}{\widthpic\textwidth}
    \includegraphics[width=\textwidth]{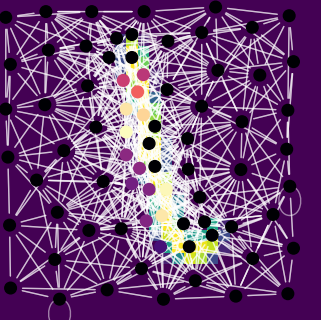}
    \caption{ScoreCAM-$n^{\circ}$478}
    \label{figvisu:SCGNN478}
  \end{subfigure}\hspace{\hsp}
  \begin{subfigure}{\widthpic\textwidth}
    \includegraphics[width=\textwidth]{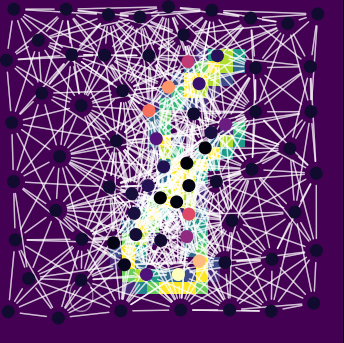}
    \caption{DeepLIFT-$n^{\circ}$444}
    \label{figvisu:DL444}
  \end{subfigure}\hspace{\hsp}%
\begin{subfigure}{\widthpic\textwidth}
    \includegraphics[width=\textwidth]{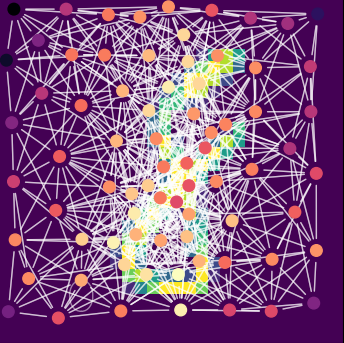}
    \caption{GradCAM-$n^{\circ}$444}
    \label{figvisu:GCAM444}
  \end{subfigure}\hspace{\hsp}%
%\begin{subfigure}{\widthpic\textwidth}
    %\includegraphics[width=\textwidth]{plt_MNIST_zkhffnbb_444_GNNEX.png}
    %\caption{\ GNNExplainer - $n^{\circ}$ 444}
    %\label{figvisu:GNNEX444}
  %\end{subfigure}\hspace{\hsp}%
\begin{subfigure}{\widthpic\textwidth}
    \includegraphics[width=\textwidth]{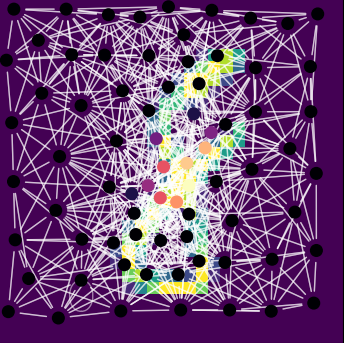}
    \caption{ScoreCAM-$n^{\circ}$444}
    \label{figvisu:SCGNN444}
  \end{subfigure}\hspace{\hsp}
  \begin{subfigure}{\widthpic\textwidth}
    \includegraphics[width=\textwidth]{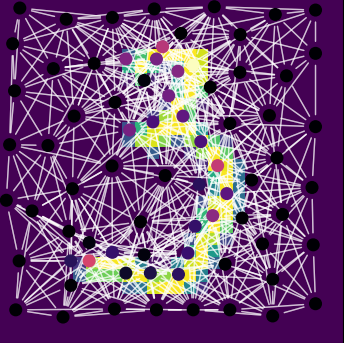}
    \caption{DeepLIFT-$n^{\circ}$969}
    \label{figvisu:DL969}
  \end{subfigure}\hspace{\hsp}%
\begin{subfigure}{\widthpic\textwidth}
    \includegraphics[width=\textwidth]{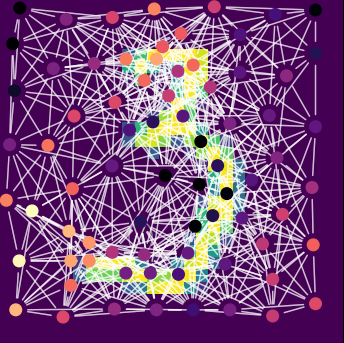}
    \caption{GradCAM-$n^{\circ}$969}
    \label{figvisu:GCAM969}
  \end{subfigure}\hspace{\hsp}%
%\begin{subfigure}{\widthpic\textwidth}
    %\includegraphics[width=\textwidth]{plt_MNIST_zkhffnbb_969_GNNEX.png}
    %\caption{\ GNNExplainer - $n^{\circ}$ 969}
    %\label{figvisu:GNNEX969}
  %\end{subfigure}\hspace{\hsp}%
\begin{subfigure}{\widthpic\textwidth}
    \includegraphics[width=\textwidth]{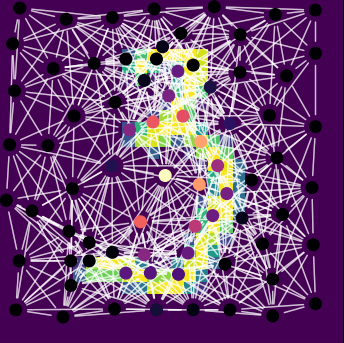}
    \caption{ScoreCAM-$n^{\circ}$969}
    \label{figvisu:SCGNN969}
  \end{subfigure}\hspace{\hsp}
\caption{Différentes explications de MNISTSuperpixels}
\label{figvisu}
\end{figure*}
On observe que notre méthode se concentre sur les descripteurs clés pour fournir ses cartes de saillance. 
\subparagraph{Pour le chiffre 1 :} Dans la Figure \ref{figvisu:SCGNN346}, la carte est distribuée spatialement de manière uniforme sur l'ensemble des nœuds superposant spatialement le chiffre considéré. En effet, la structure spatiale du chiffre est ici régulière, par conséquent aucun superpixel saillant n'a de quantité d'informations plus importante, d'un point de vue de la tâche de classification, qu’un autre. Contrairement à la méthode DeepLIFT (Figure \ref{figvisu:DL346}), l'instance produite par ScoreCAM n'introduit pas de redondance, ce qui permet une meilleure expressivité, une explication contenue et concentrée, mais conservant malgré tout ses qualités explicatives. L'instance 478 appartient à la même classe que l'instance 346. Dans la Figure \ref{figvisu:SCGNN478}, la structure du chiffre est plus complexe, pour ScoreCAM GNN, la redistribution de l'importance a été effectué majoritairement sur les points de la courbure du chiffre les plus remarquables, ceux qui qualifient le mieux la structure de celui-ci. Dans la Figure \ref{figvisu:DL478}, DeepLIFT ne parvient pas à rendre compte de l'augmentation de la complexité de la structure du chiffre et fournit une distribution équivalente à celle fournie pour l'instance 346. Elle fournit ainsi une explication équivalente alors qu'il y a bien une différence notable entre ces deux instances. Par ailleurs, DeepLIFT conserve son caractère redondant. La méthode GradCAM fournit, sur un régime constant, des cartes de saillance difficilement qualifiables et ne rendant manifestement pas compte de manière humainement intelligibles de la tâche de classification. 
\subparagraph{Pour le chiffre 8 :} L'instance 444 appartient à la classe du chiffre huit. Ce chiffre est singularisé structurellement par son point de croisement interne. Notre méthode (Figure \ref{figvisu:SCGNN444} concentre toute sa distribution de saillance sur ce point structurel remarquable qui qualifie majoritairement ce qu'est le chiffre huit. Les autres méthodes (Figure \ref{figvisu:DL444}, \ref{figvisu:GCAM444}) n'accordent pas prioritairement d'importance à cette caractéristique singulière.
\subparagraph{Pour le chiffre 3 :} L'instance 969 a une courbure importante, la carte de saillance est distribuée de manière prépondérante sur cette région, qui est objectivement caractérisante. De par sa structure, ScoreCAM construit ses cartes de saillance en respectant la construction algébrique des classifieurs sur graphes. En effet, on observe notamment Figure \ref{figvisu:SCGNN969} un centre d'importance majeur au nœud voisin des nœuds caractérisant la structure du chiffre considéré, conséquence de la phase d'agrégation additive \cite{scarselli_graph_2009}. Cette agrégation permet une synthétisation de l'interprétation de plus haut niveau de la carte de saillance. 
\section{Conclusion}
Dans cet étude, nous proposons une méthode d'explicabilité de réseaux profond sur graphes plus optimale, en étant plus fidèle au classifieur et en fournissant des cartes de saillance avec moins de redondance et plus de parcimonie tout en conservant une qualité d`explication importante. En evitant les écueils liés au conditions d'initialisation et en proposant des solutions optimales, non approximatives et ne nécéssitant pas l'optimisation d'un modèle supplémentaire, la méthode ScoreCAM surpasse les méthodes de l'art de manière objective et subjective.
\newpage
\bibliographystyle{plain}
\bibliography{gretsi22}
\end{document}